%% file: RL.tex
\documentclass{article}

\usepackage[final]{nips_2017}

\usepackage[utf8]{inputenc} 
\usepackage[T1]{fontenc}    
\usepackage{url}            
\usepackage{booktabs}       
\usepackage{amsfonts}       
\usepackage{nicefrac}       
\usepackage{microtype}      
\usepackage{mathrsfs,amsmath}
\usepackage{multirow}

\usepackage{graphicx}
\usepackage[hidelinks]{hyperref}

\title{Reinforcement Learning To Adapt Speech Enhancement to Instantaneous Input Signal Quality}

\author{
   Rasool Fakoor \thanks{Work was done as an intern at Microsoft AI and Research, Redmond. \newline  Correspondence to:
    Rasool Fakoor <rasool.fakoor@mavs.uta.edu>.} \\
   Dept. of Computer Science and Engineering\\
   Univ. of Texas at Arlington, TX 76019\\
   \texttt{rasool.fakoor@mavs.uta.edu} \\
   \And
   Xiaodong He, Ivan Tashev and Shuayb Zarar \\
   Microsoft AI and Research \\ Redmond WA 98052 \\
   \texttt{{xiaohe,ivantash,shuayb}@microsoft.com} \\
}

\begin{document}

\maketitle

\input{Abstract}
\input{Introduction}

\input{Model}

\input{Experiments}
\input{Conclusion}

\bibliographystyle{apalike}
\bibliography{refs}

\end{document}

%% file: Abstract.tex
\begin{abstract}
Today, the optimal performance of existing noise-suppression algorithms, both data-driven and those based on classic statistical methods, is range bound to specific levels of instantaneous input signal-to-noise ratios. In this paper, we present a new approach to improve the adaptivity of such algorithms enabling them to perform robustly across a wide range of input signal and noise types. Our methodology is based on the dynamic control of algorithmic parameters \textit{via} reinforcement learning. Specifically, we model the noise-suppression module as a black box, requiring no knowledge of the algorithmic mechanics except a simple feedback from the output. We utilize this feedback as the reward signal for a reinforcement-learning agent that learns a policy to adapt the algorithmic parameters for every incoming audio frame (16 ms of data). Our preliminary results show that such a control mechanism can substantially increase the overall performance of the underlying noise-suppression algorithm; 42\% and 16\% improvements in output SNR and MSE, respectively, when compared to no adaptivity.
\end{abstract}

%% file: Introduction.tex
\section{Introduction}\label{sec:intro}
Noise-suppression algorithms for a single-channel of audio data employ machine-learning or statistical methods based on the amplitude of the short-term Fourier Transform of the input signal \citep{Ephraim:84, ephraim1995signal, boll1979suppression, xu2014experimental}. Although the approach we propose can be applied to the entire gamut of noise-suppression techniques, in this paper, we only illustrate its benefits with the classical algorithms for speech enhancement that are based on spectral restoration~\citep{Tashev2009}. Such algorithms typically comprise four components~\citep{Tashev2009}: (1) voice-activity detection, (2) noise-variance estimation, (3) suppression rule, and (4) signal amplification. The first two components help gather statistics on the target speech signal in the input audio, while the third and fourth components allow us to utilize these statistics to distill out the estimated speech signal. Despite these components being based on sound mathematical principles~\citep{Tashev2009}, their performance is directly influenced by a sizeable set of parameters such as those that control the gain, geometry weighting, estimator bias, voice- and noise-energy thresholds, and \textit{etc}. Consequently, the combined set of these parameters plays a critical role in achieving the best performance of the end-to-end speech-enhancement process. Needless to say, the numerical values of these parameters are heavily influenced by the input signal and noise characteristics. Furthermore, thanks to the complex interdependency between statistical models~\citep{Tashev2009}, there is no known best value for these parameters that works well across all levels of input signal quality. Therefore any offline optimization process, such as the simplex method~\citep{Nash20020}, is only a \textit{sub-optimal solution} that tends to achieve good performance across only a small range of instantaneous input signal-to-noise ratios (SNRs). This is the status quo that we intend to break.

%% file: Model.tex
\section{Proposed Approach}\label{sec:model}

We develop data-driven techniques that allow us to adjust control parameters of a classical speech-enhancement algorithm~\citep{Tashev2009}  dynamically at a frame level, depending just on simple feedback from the underlying algorithm. Specifically, we rely on reinforcement-learning (RL) based on a network of long-short term memory (LSTM) cells \citep{Hochreiter}. We show that our method can achieve the best performance of speech enhancement across a broad range of input SNRs.

\begin{figure}[tb]
\centering
\centerline{\includegraphics[width=0.7\textwidth]{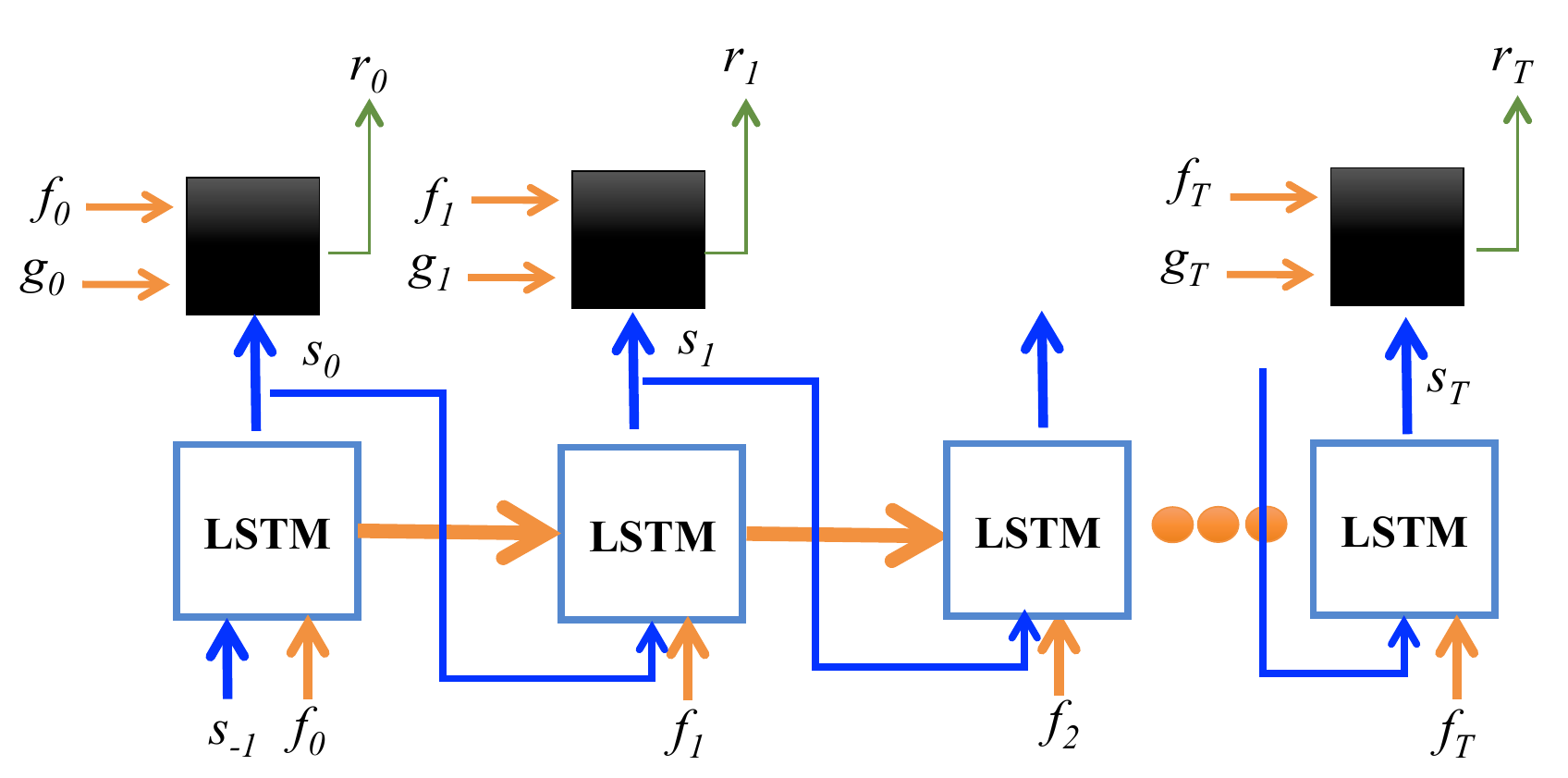}}
\caption{Proposed architecture. $g_t$ and $f_t$ are clean and noisy input frames, while $s_t$ and $r_t$ are action and reward values at time step $t$, respectively. At each time step, our model utilizes $s_t$ and $f_t$ to find the best set of control parameters that maximize $r_t$ of the speech-enhancement black-box algorithm.}
\label{fig:ourmodel}
\end{figure}

The neural-network model that we propose is shown in Fig.~\ref{fig:ourmodel}. As mentioned before, it treats the classical speech-enhancement algorithm~\citep{Tashev2009} as a black box. Suppose $s_t$ and $r_t$ represent the action proposed by our network model and the reward signal that is available to it from the algorithm at an instance of time $t$. We set up an objective function as follows:
\begin{equation}
\label{eq:objective_function}
J^\pi(\theta) = \mathbb{E}_{\pi_\theta} \bigg[\sum_{t=0}^T r_t(s_t)\bigg]
\end{equation}
where $\pi_{\theta}$ is the policy that our model tries to learn. The goal of this function is to maximize the expected reward over time. Thus, in order to solve Eq.~\eqref{eq:objective_function}, we employ the REINFORCE algorithm \citep{Williams, ZarembaS15, RanzatoCAZ15, NIPS2014_5542,baVald2015,s2sIlya}. At each time step, our network picks an action   at a given state of the model (i.e. given it policy $\pi_{\theta}$) that causes some change to the set of control parameters that are applied to the black box. In other words, each parameter of the speech-enhancement algorithm~\citep{Tashev2009} is mapped to an action. Thus, an action can result in an increase or decrease of the parameter value by a specific step size. It could also lead to no change in the parameter values. Note that within the black-box algorithm, we also utilize the clean signal $g_t$ in addition to the noisy frame at every time step. The reason we do this is because once the underlying speech-enhancement algorithm~\citep{Tashev2009} is done with the denoising process, it relies on the ground-truth clean signal $g_t$ to compute a score that represents the reward function for the RL agent. $g_t$ is not used within the black box in any other way.

%% file: Experiments.tex
\section{Experimental Results}\label{sec:con}

We evaluated the performance of our methodology with single-channel recordings based on real user queries to the Microsoft Cortana Voice Assistant. 
We split studio-level clean recordings into training, validation and test sets comprising $7500$, $1500$ and $1500$ queries, respectively. Further, we mixed these clean recordings with noise data (collected from 25 different real-world environments), while accounting for distortions due to room characteristics and distances from the microphone. Thus, we convolved the resulting noisy recordings with specific room-impulse responses, and scaled them to achieve a wide input SNR range of 0-30 dB. Each (clean and noisy) query has on average more than $4500$ audio frames of spectral amplitudes, each lasting $16$ ms. We applied a Hann weighting window to the frames allowing accurate reconstruction with a $50\%$ overlap. These audio frames in the spectral domain formed the features for our algorithm. Since we utilized a $512$-point short-time Fourier Transform (STFT), each feature vector was a positive real number of dimensionality $256$. To train our network model, we employed a first-order stochastic gradient-based optimization method, Adam ~\citep{KingmaB14}, with a learning rate that was adjusted on the validation set. Furthermore, we used a single layer LSTM  with $196$ hidden units and number of steps equal to the number of frame. We trained it with a batch size of $1$ given each of input file has different number of frames.

To compute the reward function, we employed the negated mean-squared error (MSE) between the ground-truth clean signals $g_t$ and denoised input signals $\hat{g}_t$ as follows:
%
\begin{equation}
\label{eq:reward}
r_t = - \Vert g_t - \hat{g}_t \Vert^2_2.
\end{equation}
%
Further, to avoid instability during training, we normalized this reward function to lie between $[-1,1]$. Our underlying speech-enhancement algorithm~\citep{Tashev2009}, i.e black-box, was based on a generalization of the decision-directed approach, first defined in \citep{Ephraim:84}. To quantify the performance of speech enhancement, we employed the following metrics:
%
\begin{itemize} 
 \setlength{\itemsep}{-1pt}
 \item Signal-to-noise ratio (SNR) dB
 \item Log spectral distance (LSD)
 \item Mean squared error in time domain (MSE)
 \item Word error rate (WER)
 \item Sentence error rate (SER)
 \item Perceptual evaluation of speech quality (PESQ)
\end{itemize}
%
%
\begin{table}[t]
  \centering
  \begin{tabular}{cllllll}
    \toprule
     Method & {\sc SNR}(dB) & {\sc LSD} &{\sc MSE}& {\sc WER} & {\sc SER} & {\sc PESQ}\\ 
    \hline \\ [-3pt]
    \textbf{Noisy Data}      & $15.18$ & $23.07$ & $0.04399$ & $15.4$ & $25.07$ & $2.26$ \\ [3pt]
    \textbf{Baseline~\citep{Tashev2009}} & $18.82$ & $22.24$ & $0.03985$ & $14.77$ & $25.93$ & $2.40$ \\ [2pt]
    \textbf{Proposed (Unbiased Estimation)}      & $\mathbf{26.16}$ & $\mathbf{21.48}$ & $\mathbf{0.03749}$ & $17.38$ & $31.87$ & $2.40$ \\ [2pt]
	\textbf{Proposed (Baselined Estimation)} & \multirow{1}{*}{$\mathbf{26.68}$} & \multirow{1}{*}{$\mathbf{21.12}$} & \multirow{1}{*}{$\mathbf{0.03756}$} & \multirow{1}{*}{$18.97$} & \multirow{1}{*}{$32.73$} & \multirow{1}{*}{$2.38$} \\ [3pt]
    \textbf{Clean Data}      & $57.31$ & $1.01$ & $0.0$ & $2.19$ & $7.4$ & $4.48$ \\
    \bottomrule
  \end{tabular}
  \vspace*{10pt}
\caption{The proposed RL approach improves MSE, LSD and SNR with no algorithmic changes to the baseline speech-enhancement process, except frame-level adjustment of the control parameters.}
 \label{tb:results}
  \vspace*{-5pt}
\end{table}
%
A larger value is desirable for the first and last metrics, while a lower value is better for the rest. To compute the WER and SER, we employed a production-level automatic speech recognition (ASR) algorithm, whose acoustical model was trained separately on a different dataset that had similar statistics as our training examples. Thus, the ASR algorithm was not re-trained during our speech-enhancement experiments. Results of evaluating our model on the test data are shown in Table~\ref{tb:results}. In the baseline approach~\citep{Tashev2009}, we utilized a non-linear solver to find the set of algorithmic parameters that achieved the best score for a multi-variable function that equally weighted all of the above metrics. This unconstrained optimization was performed \textit{offline} once across the training data, which resulted in a parameter set that achieved the best trade-off for all metrics and feature vectors across the input SNR range. This parameter set was held constant when the speech-enhancement algorithm~\citep{Tashev2009} was applied to the test audio frames. However, in the proposed approaches (third and fourth rows in the table), the parameter set was adjusted depending on the action proposed by our RL meta-network. The third row corresponds to utilizing LSTM-based RL alone on top of the baseline speech-enhancement algorithm [also known as the unbiased estimator that utilizes the reward function of Eq.~\eqref{eq:reward}]. While in the fourth row, we add an additional step of reducing the variance of gradient estimation by baselining Eq.~\eqref{eq:reward}. 

From Table~\ref{tb:results}, we see that the proposed models show better performance on MSE, LSD and SNR, improving them by up to $16\%$, $4\%$, and $42\%$, respectively. However, they do not show improvement on the other metrics (WER, SER and PESQ). In fact, these results are expected because our RL network only employs a measure for signal distortion as the reward function [see Eq.~\eqref{eq:reward}]. Thus, it is able to only optimize metrics that are related to this measure (\textit{i.e.}, MSE, SNR and LSD). The difficulty of including WER, SER and PESQ in the RL optimization process lies in the fact that these metrics do not provide a direct way of quantifying representation error. There is also no good signal-level proxy for them that can be computed with a low processing cost, which is necessary to train the RL algorithm in practical amounts of time. Thus, although our network already deals with a hard and complex optimization problem due to the black-box optimization and policy gradients, as part of future work, we are continuing to investigate different methods of incorporating WER, SER and PESQ functions into the RL reward signal.

%% file: Conclusion.tex
\section{Discussion and Future Work}\label{sec:conFuture}

In this ongoing work, we proposed a method based on black-box optimization and RL to deal with the problem of adapting speech-enhancement algorithms to varying input signal quality. Our work is related to hyperparameter optimization in deep learning and machine learning\footnote{Here, hyperparameters refer to training parameters such as learning rate, decay function or a network structure such as the number of layers, number of hidden units, type of layers \textit{etc.}}, which has been extensively studied in the literature. Methods like random search~\citep{Bergstra2012}, Bayesian optimization with probabilistic surrogates (\textit{e.g.}, Gaussian processes~\citep{Snoek2012, HenrandezLobato2014}) or deterministic surrogates (\textit{e.g.}, radial basis functions~\citep{ilievski2017efficient}) have been used to find the best setup for the model hyperparameters. However, once these methods find a set of parameters for a given model offline, the set typically remains fixed throughout the inference process. In contrast, our RL approach adaptively changes parameters of the underlying (data-driven or analytical) algorithm at inference time, achieving the best performance under all input signal conditions. Furthermore, in this particular paper, we demonstrated how to apply our dynamic parameter-adaptation technique to the problem of speech enhancement~\citep{Tashev2009}. To the best of our knowledge, black-box optimization using reinforcement learning for real-time application such as speech enhancement has not been conducted before, the previous work~\citep{chen17eLTLWGD} only studies a simple synthetic task. Based on experiments with real user data, we showed that our RL agent is very effective in changing algorithmic parameters at a frame level, enabling existing speech-enhancement algorithms to adapt to changing input signal quality and denoising performance. However, there are still hurdles that need to be overcome in the design of a reliable reward function that helps us achieve the best algorithmic performance across a diverse range of metrics including WER, SER and PESQ. We intend to address this challenge in future work, in addition to reducing the overhead of RL computation during training.